%% file: nips2016snnNoFB.tex
\documentclass[10pt,a4paper]{article}
\usepackage[normalem]{ulem}
\usepackage{url}  
\usepackage[T1]{fontenc}
\usepackage{booktabs}
\usepackage[english]{babel}
\usepackage[nohead,includefoot,margin=2.5cm]{geometry}
\usepackage[compact,raggedright,small]{titlesec}
\usepackage[affil-it]{authblk}
\usepackage[runin]{abstract}
\usepackage{multicol}
\usepackage{microtype} 
\usepackage{nicefrac} 
\usepackage{amsfonts}

\usepackage[utf8]{inputenc} 
\usepackage[table]{xcolor}
\usepackage{amsmath}
\usepackage{graphicx}
\usepackage{textcomp}
\usepackage{wrapfig}
\usepackage{multirow}
\usepackage{icomma} 
\usepackage{subcaption}

\usepackage{lipsum}

\let\OLDthebibliography\thebibliography
\renewcommand\thebibliography[1]{
  \OLDthebibliography{#1}
  \setlength{\parskip}{4pt plus 1.3ex}
  \setlength{\itemsep}{0pt plus 1.3ex}
}

\title{\large\bfseries Fast and Efficient Asynchronous Neural Computation with Adapting Spiking Neural Networks}
\date{\small \today}
\author[1]{Davide Zambrano }
\author[1]{Sander M. Bohte}
\affil[1]{Machine Learning group, Centrum Wiskunde \& Informatica (CWI), 1098XG Amsterdam, The Netherlands}

\begin{document}

\maketitle

\input{intro}

\input{multadaspi}

\input{methodsondataset}

\input{resultsffnn}

\input{discuss}

\bibliographystyle{unsrt}
\bibliography{SNN}

\end{document}

%% file: intro.tex
\begin{abstract}
Biological neurons communicate with a sparing exchange of pulses -- spikes.
It is an open question how real spiking neurons produce the kind of powerful neural computation that is possible with deep artificial neural networks, using only so very few spikes to communicate. Building on recent insights in neuroscience, we present an Adapting Spiking Neural Network (ASNN) based on adaptive spiking neurons. These spiking neurons efficiently encode information in spike-trains using a form of Asynchronous Pulsed Sigma-Delta coding while homeostatically optimizing their firing rate. In the proposed paradigm of spiking neuron computation, neural adaptation is tightly coupled to synaptic plasticity, to ensure that downstream neurons can correctly decode upstream spiking neurons. We show that this type of network is inherently able to carry out asynchronous and event-driven neural computation, while performing identical to corresponding artificial neural networks (ANNs). In particular, we show that these adaptive spiking neurons can be drop in replacements for ReLU neurons in standard feedforward ANNs comprised of such units. We demonstrate that this can also be successfully applied to a ReLU based deep convolutional neural network for classifying the MNIST dataset. The ASNN thus outperforms current Spiking Neural Networks (SNNs) implementations, while responding (up to) an order of magnitude faster and using an order of magnitude fewer spikes. Additionally, in a streaming setting where frames are continuously classified, we show that the ASNN requires substantially fewer network updates as compared to the corresponding ANN.

\end{abstract}
\section{Introduction}

With rapid advances in deep neural networks, renewed consideration is given to the question how artificial neural networks relate to the details of information processing in real biological \emph{spiking} neurons. Apart from its still vastly more flexible operation, the huge spiking neural network that comprises the brain intrinsically operates in an asynchronous manner and is highly energy efficient. These properties derive in large part from the brain's sparse spiking activity: spikes are only emitted when a spiking neuron is sufficiently stimulated, and otherwise the neuron remains silent. Intrinsic neural mechanisms also homeostatically control a neuron's firing rate to optimally encode information \cite{fairhall2001efficiency}. As a result, estimates are that neurons in mammalian brains on average only emit somewhere between 0.2--5 spikes per second \cite{attwell2001energy}. 

From an AI perspective, low and event-activated neural networks are attractive for applications with always-on requirements, for instance for use in cell-phones. Neuromorphic implementations of spiking neural networks have been specifically developed to create energy efficient implementations of deep neural networks \cite{merolla2014million,esser2016convolutional}. However, current spiking neural networks that almost match standard deep neural network performance require the use of very high firing rates, negating much of the improved efficiency. 

\begin{figure}
\begin{center}
\includegraphics[width=0.99\linewidth]{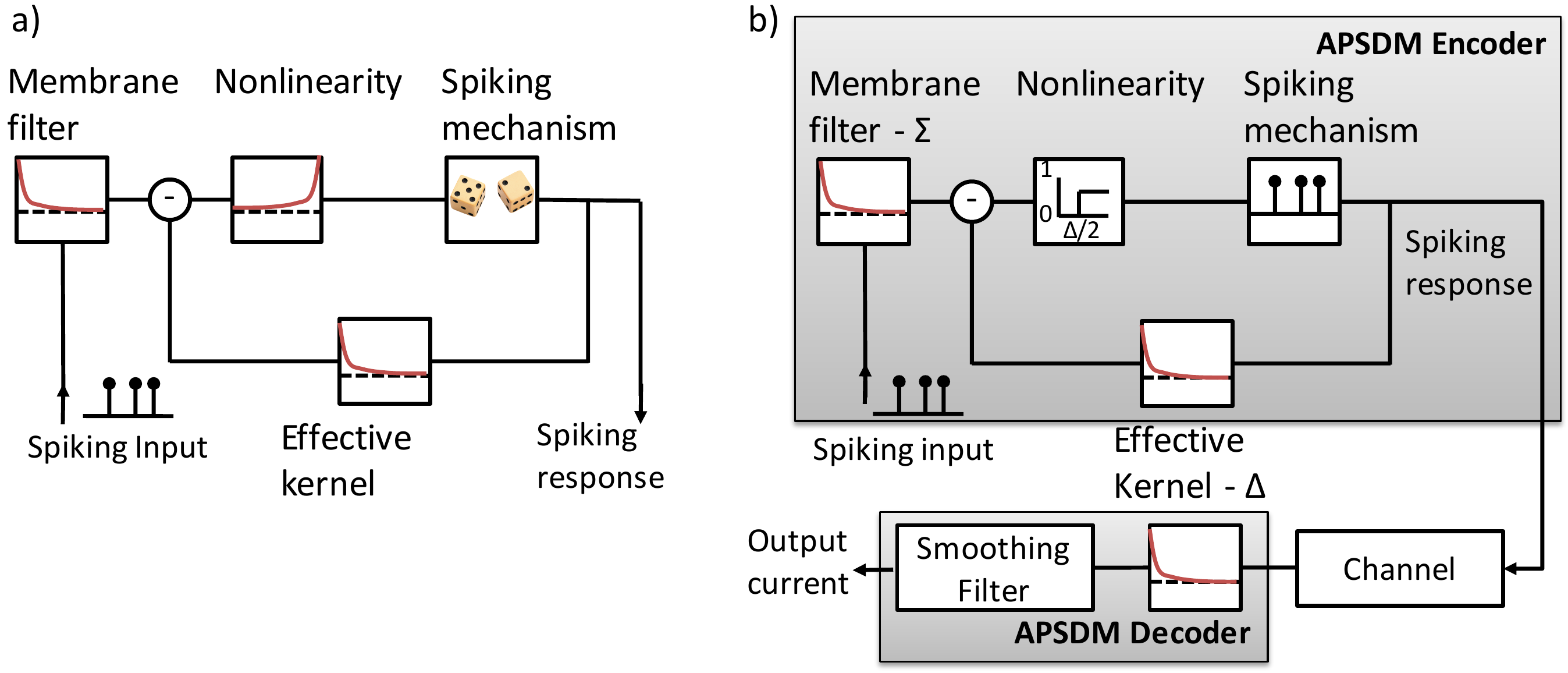}
\end{center}
\caption{\small (a) Generalized Leaky Integrate-and-Fire model and (b) the Asynchronous-Pulse-Sigma-Delta-Modulation (APSDM) \cite{Yoon:hv} (right). Note the close similarities.}
\label{fig:GLIFandAPSDM}
\end{figure}

It is still an open question in neuroscience how exactly biological spiking neurons convey information to each other using as few spikes as possible \cite{rieke1999spikes,GerKis02,deneve2016efficiency}. Neural units in a standard Artificial Neural Network (ANN) propagate analog values to downstream neurons whereas biological spiking neurons emit spikes that, through neurotransmitters released at the synaptic connection, exert influence on the state of the target neuron. In the standard firing-rate formulation, spiking neurons are presumed to emit isomorphic spikes according to an inhomogenous Poisson process with a rate proportional to the internally computed analog value (i.e. Figure \ref{fig:GLIFandAPSDM}a): the spiking neuron's non-linear average response approximates a ReLU neurons in neural network terms over some limited dynamic range. 

As was shown by Liu et al \cite{oconnor2013real,diehlIJCNN2015,neil2016learning}, with some work the ReLU neuron in standard neural networks and convolutional neural networks can be replaced by Poisson spiking neurons; this approach obtained performance close to the corresponding ANN and is to our knowledge the currently best performing spiking neural network. Still, this spiking neural network required very high firing rates (200-1000Hz) and substantial temporal averaging (150-500ms) to obtain accurate-enough estimates of firing rate (and hence classification). To obtain faster responses the weights obtained in the ANN model had to be reweighted and/or normalised to account for the limited dynamic range of simple Poisson neurons. Similarly, Cao et al \cite{Cao2015ijcv} create deep spiking neural networks based on LIF neurons that use spike-counting for classification over a 100, 200 or 400ms interval from the start of input presentation. 

To explain how real neurons are able to efficiently encode a signal with few spikes, alternative spike-based neural coding schemes are being considered. A recent line of work in computational neuroscience suggests that spiking neurons may implement a form of online AD/DA conversion \cite{bohte_nips2010,deneve2011,Bohte:2012tf,Chklovskii2012NIPS,Yoon:hv,deneve2016efficiency}; the key idea is that when a neuron spikes, the refractory reset removes a part of the internally computed analog voltage signal. The resultant spike, through the synapse, then delivers this quantum of signal to the next neuron. Yoon \cite{Yoon:hv} recently demonstrated a direct correspondence between Generalized Leaky-Integrate-and-Fire (G-LIF) models (ie \cite{Pozzorini:2013bj}) and the AD/DA encoding/decoding scheme called Asynchronous-Pulse-Sigma-Delta Modulation (APSDM) developed for digital signal processing (compare also Figure \ref{fig:GLIFandAPSDM}a and b). The APSDM scheme however presumes a fixed dynamic range for the encoded analog values and still requires very high firing rates for good AD/DA signal approximations. 

In this paper, we build on these recent insights to develop a novel kind of asynchronous {\em adapting spiking neural network} that is highly efficient in terms of spikes used, based on adaptive spiking neurons that can function as drop-in replacements for standard ReLU neurons in traditional (deep) neural networks. Inspired by \cite{brette2011spiking} and \cite{Bohte:2012tf}, we use a multiplicative model of neural adaptation to obtain a spiking ReLU neuron that is capable of encoding and decoding a wide dynamic range of activation values with a limited and tunable firing rate. We thus effectively extend the APSDM scheme with a (biologically plausible) spiking neuron model that dynamically adapts its spiking mechanism to the (varying) dynamic range of the computed internal activation value. From a biological perspective, the proposed adaptive neural coding model predicts that various forms of somatic adaptation operate in a concerted manner with synaptic phenomena like short term synaptic plasticity, possibly combined with effects like synaptic scaling. 

We show that we can thus compose computationally efficient Adaptive Spiking Neural Networks (ASNNs) through drop-in replacement of ReLU neurons in standard ReLU-based feedforward and convolutional ANNs, and we demonstrate identical performance to these ANNs without additional modifications. The ASNNs outperform previous SNNs like \cite{diehlIJCNN2015,neil2016learning} on a selection of standard benchmarks, including MNIST; require an order of magnitude fewer spikes, albeit analog rather than binary spikes (and still more efficient in terms of communicated bits); and are also up to an order of magnitude faster, in the sense that they need much less temporal averaging over output spike-trains. Additionally, the presented network is able to carry out ongoing asynchronous neural computation in continuous time: neurons are updated at high temporal precision and information is exchanged only sparingly in an asynchronous manner. This has to be compared to the traditional ANN which compute in a fully synchronous fashion: in a streaming classification problem, we show that for ASNNs a much sparser network updating suffices as compared to traditional ANNs.  

The novel type of neural network described here strongly leans on neural coding principles suggested from neuroscience to arrive at a formulation of neural computation that falls in between classical ANNs and SNNs. The work described here offers a new and essentially hybrid paradigm of low firing rate spiking neural networks. 
    
%% file: multadaspi.tex
\section{Multiplicative adaptive spike-time coding}

\begin{figure}
  \centering
  \includegraphics[width=0.99\linewidth]{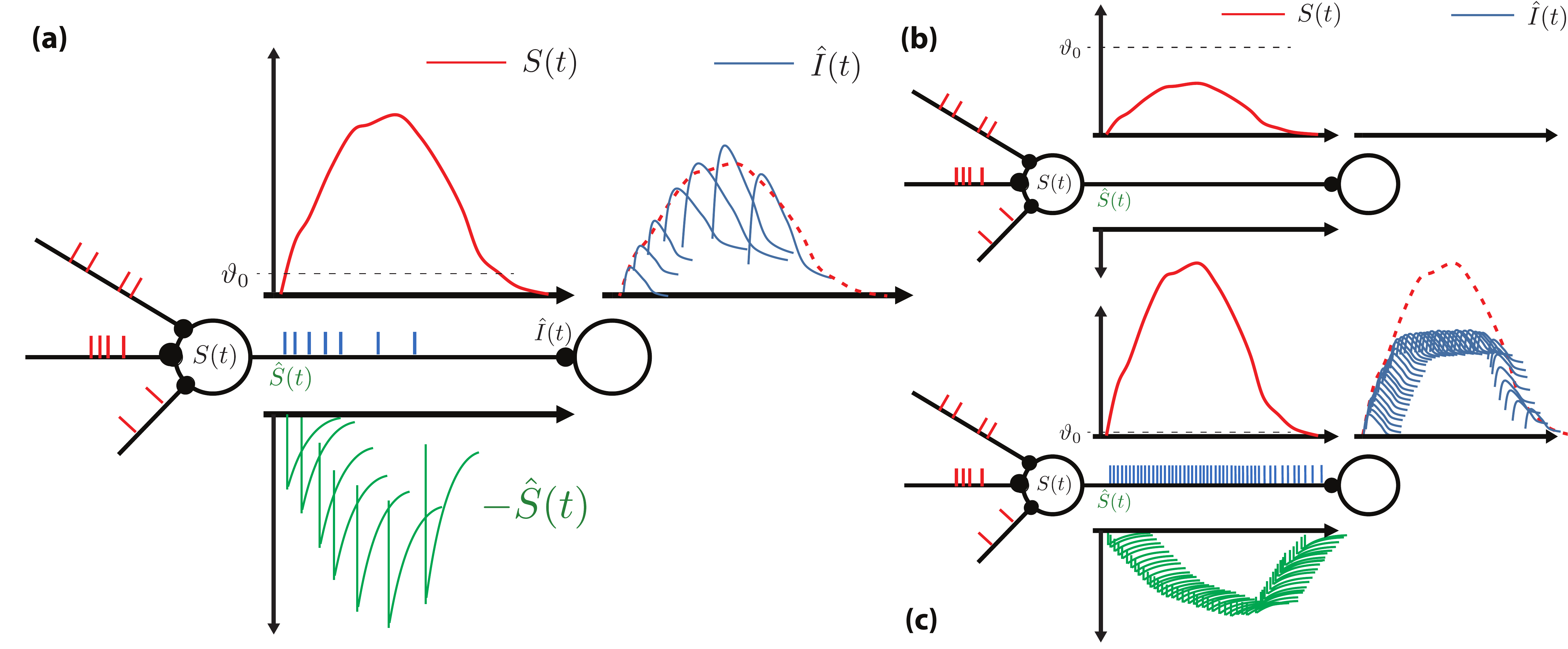}
 \caption{\small (a) Illustration of signal encoding with the ASN. $\hat{I}$ denotes the smoothed sum of (weighted) postsynaptic currents in the post-synaptic target neuron, proportionally approximating the encoded presynaptic signal $S(t)$. (b,c) Limited dynamic range: approximations fail when the signal $S(t)$ is too small relative to the neurons' threshold $\vartheta_0$ (no spikes), or, (c) too large: then, due to absolute refractoriness and corresponding maximum firing rate, the ``high'' parts of the signal $S(t)$ cannot be encoded.}
\label{fig:ASneuron}
\end{figure}

A spiking neural network is defined by the relationship between spikes and the quantity that is computed in the neuron as the result of impinging spikes \cite{maass1997networks}. We define \emph{sigma-delta spike-time coding} as the neural equivalent of the APSDM framework (Figure \ref{fig:GLIFandAPSDM}b): weighted input spikes contribute linearly to the membrane potential, and when this sum of inputs reaches the threshold from below, a spike is generated and a refractory reset is subtracted from the membrane potential. When both refractory reset and the impact of spikes on downstream neurons is proportional and temporally extended, effectively it is the (smoothed) sum of refractory resets that is conveyed to the next neuron; this AD/DA encoding/decoding process is illustrated in Figure \ref{fig:ASneuron}a: the input spikes (red) collectively cause an input voltage $S(t)$ (red line). Due to thresholding, this input voltage causes a series of spikes to be emitted, and each spike-generation subtracts a refractory response from the incoming potential (green), the sum of which, $\hat{S}(t)$ approximates the input voltage $S(t)$. At the receiving neuron, each spike triggers a post-synaptic potential proportional to the refractory response, weighted by the size of the synaptic weight, and smoothed due to the membrane resistance of the postsynaptic neuron. With only a single input and an appropriately chosen threshold $\vartheta_0$, this smoothed sum of postsynaptic potentials, $I(t)$ proportionally approximates the shape of the signal computed in the pre-synaptic neuron, $S(t)$ (red dotted line). In this scheme, only positive parts of $S(t)$ are approximated, thus effectively computing a ReLU function on the presynaptic's neuron input. 

\paragraph{An Adaptive Spiking Neuron}

To create artificial spiking neural networks based on sigma-delta spike-time coding, we address the limited dynamic range of standard LIF or corresponding Spike-Response-Model (SRM) neurons. Here, we note that it is the fixed size refractory resets that limit the dynamic range of the internal activation that a neuron can encode with the proposed spike-time coding mechanism \cite{Bohte:2012tf,Chklovskii2012NIPS}. Effectively, activation values that are either too small or too large relative to the threshold cannot be accurately encoded. This is illustrated in figure \ref{fig:ASneuron}(b,c). 

Here we use the solution proposed in \cite{Bohte:2012tf} based on a model of fast adaptation in spiking neurons: by dynamically adjusting the threshold, the size of the refractory responses can be controlled and the dynamic range can be increased, drastically even when a multiplicative form of threshold adjustment is used. Such multiplicative adaptation effectively allows a neuron to assign a fixed ``budget'' of spikes to a given dynamic range, also when that range changes drastically. Note that such a model of adaptation explains various adaptive behaviour in real biological neurons \cite{fairhall2001efficiency,brette2011spiking,Bohte:2012tf}. 
 
We implement {\em adaptive} sigma-delta spike-time coding using multiplicative adaptation in an SRM \cite{GerKis02}. A spiking neuron computes a smoothed internal activation value $S(t)$ on the input current:\vspace{-0.1cm}

\[
S(t) = (\phi \ast I)(t), 
\]
where $\phi(t)$ is the (exponential) smoothing filter with time constant $\tau_{smooth}$ and $I(t)$ is the input current that the neuron receives. This current $I(t)$ can be injected directly into the spiking neuron (for inputs), or be the result of impinging (weighted) spikes causing post-synaptic currents (PSCs) (specified below). 
The spiking mechanism approximates the ReLU activation of $S(t)$ with $\hat{S}(t)$ using a sum of spike-triggered kernels $\eta(t-t_i)$: \vspace{-0.1cm}
\begin{equation}\label{eq:approx}
\hat{S}(t) = \sum_{t_i} \eta(t-t_{i}),
\end{equation}
where a spike is added in an online and incremental fashion when the difference between the input signal and the signal approximation exceeds a positive dynamic threshold $\vartheta(t)$ from below:
\begin{equation}
u(t) = S(t) - \hat{S}(t) > \vartheta(t),
\end{equation}
where we take $u(t)$ to denote the neuron's membrane potential. Upon emitting a spike at $t_i$, the spike-triggered refractory response $\eta(t-t_i)$ is subtracted from $S(t)$ and added to $\hat{S}(t)$. The part of $S(t)$ larger than the minimal value of the threshold $\vartheta(t)$ is thus encoded as $\hat{S}(t)$ in a spike-train $t_i$. It is decoded at the postsynaptic target neuron where the resultant postsynaptic currents (PSCs) are added as weighed versions of the refractory response $\eta(t)$. The resultant postsynaptic current in target neuron $j$, $I_j(t)$ induced by presynaptic spikes $t_i$ from multiple presynaptic neurons $i$, is then computed as:
\[
I_j(t) = \sum_{i} \sum_{t_i} w_{ij} \eta(t-t_i),
\]
where $w_{ij}$ is the weight between presynaptic neuron $i$ and postsynaptic neuron $j$. The refractory response kernel $\eta(t)$ is adaptive and controlled through the dynamic threshold $\vartheta(t)$:
\[
\eta(t-t_i) = \vartheta(t_i) \nu(ISI) \kappa(t-t_i),
\]
where $\vartheta(t_i)$ is the effective threshold at the time of spiking, $\kappa(t-t_i)$ is a spike-triggered kernel shaping the refractory response due to the spike at $t_i$ and is computed as an exponential: $\kappa(t) =  \exp(-(t)/\tau_{\kappa})$. The factor $\nu(ISI)$ is a function of the interspike interval (ISI) between current spike $t_i$ and previous spike $t_{i-1}$ and corrects for the fact that the mean value of an $\eta$ kernel between two spikes is not half the height of $\vartheta(t_i) \kappa(t-t_i)$:
\begin{align}
\nu(t_i - t_{i-1}) &= \frac{(t_i-t_{i-1})} {2 \int_{t_{i-1}}^{t_i}  \kappa(t-t_{i-1}) dt}.
\end{align}
Thus computed, the average of the sum of $\eta$ kernels approximates the mean of the signal $S(t)$. We approximate this function with a simple linear function $\nu(ISI) = a + b\cdot ISI$ from numerical simulations.

We model the dynamic threshold $\vartheta(t)$ as multiplicative adaptation after \cite{Bohte:2012tf}: 
\begin{align}
\vartheta(t) &= \vartheta_0 + \sum_{t_i} m_f \vartheta(t_i)\gamma(t-t_i),
\end{align}
where $\vartheta_0$ is the baseline threshold set to some (small) fixed value. A multiplicative factor $m_f$ of fixed size regulates the threshold dynamics, where the ratio between $\vartheta_0$ and $m_f$ determines the asymptotic firing rate of the neuron for large activation values. The adaptation kernel $\gamma(t)$ is computed as an exponential: $\gamma(t) = \exp(-(t)/\tau_{\gamma})$. 

The Adaptive Spiking Neuron (ASN) thus defined in terms of a Spike Response Model corresponds to a variant of the well-known Generalized Leaky-Integrate-and-Fire (G-LIF) neuron \cite{GerKis02,Pozzorini:2013bj}. The neuron state update can thus be efficiently computed by updating these exponential functions as simple (memory-less) dynamical systems. 

As noted, the signal approximation $\hat{S}(t)$ is computed as a sum of variable height kernels: it is this signal that is communicated through a sequence of spikes to the next, postsynaptic, neuron. At the postsynaptic neuron, the filter $\phi(t)$ smoothes the (weighted) $\eta$ kernels, which suppresses high frequency noise and reconstructs the signal as in the APSDM receiver \cite{Yoon:hv}. In the network, for each arriving spike the corresponding $\eta$ kernel is multiplied by the weight of the connection and added to the current $I(t)$ in the post-synaptic neuron. Since the height of the $\eta$ kernel is adaptive, in this treatment each spike $t_i$ effectively has a {\em height} 
$\vartheta(t_i)$. Conceptually, the synapse converts the binary spikes into the variable and weighted contribution to the post-synaptic neuron, using the same adaptation-driving spike history effects to compute the approximate effective impact: the DA part of the AD/DA conversion. Whether binary spikes are communicated and the D/A value is computed, or whether just analog spikes are communicated is effectively a tradeoff between computation and bandwidth. In our implementation, the ASN communicates spikes with an analog ``height'' rather than binary valued spikes.

\begin{figure}[t]
\begin{center}
\includegraphics[width=0.99\linewidth]{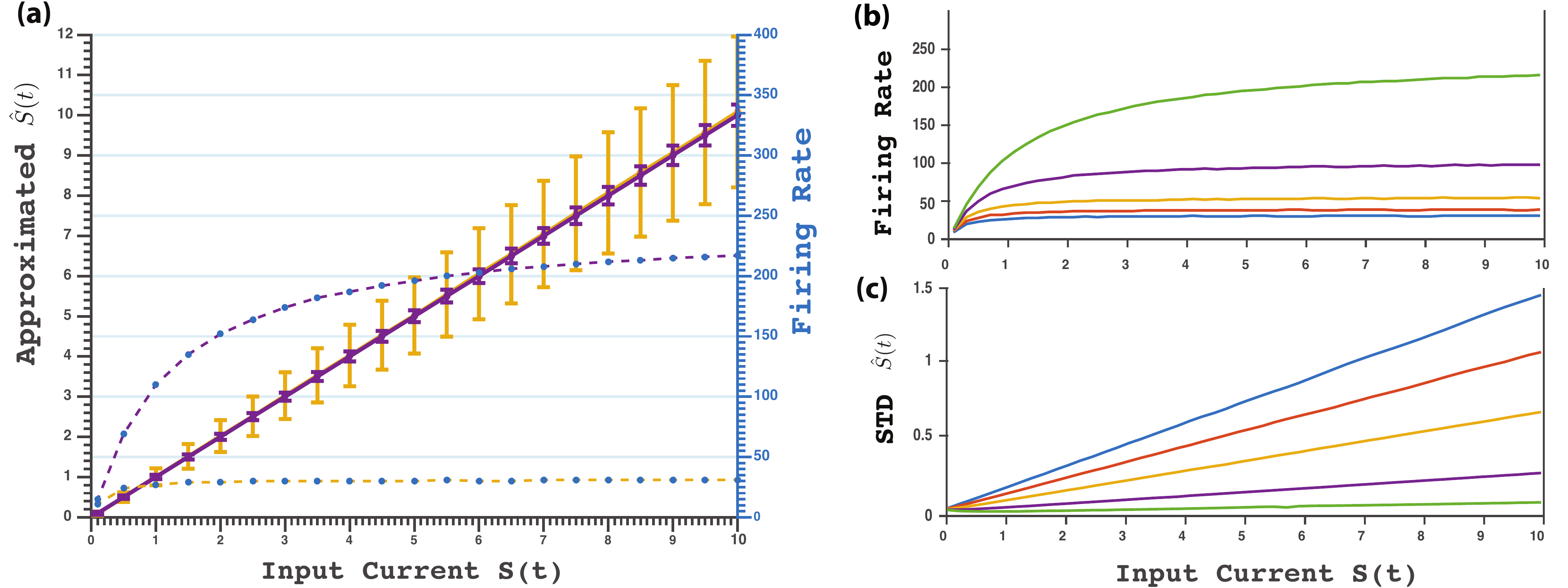}
\end{center}
\caption{\small (a) Firing rates (dashed lines, right axis, computed over a 1s time window), output signal $\hat{S}(t)$ with standard deviation (solid lines, left axis) of an ASN  ReLU neuron for two firing rate regimes ($\vartheta_0 = 0.1$, $m_f= \vartheta_0$ (yellow), $m_f= 0.1\vartheta_0$ (purple)). Colors are the same for firing rate and corresponding signal $\hat{S}$. (b) Firing rate for 5 different values of $m_f = 0.01, 0.025, 0.05, 0.075, 0.1$ and $\vartheta_0=0.1$. (c) Standard deviation (std) for 5 different values of $m_f$. Colors correspond between (b) and (c).}
\label{fig:ReLuVariance}
\end{figure}

\paragraph{Adaptive Signal Encoding and Decoding}

The signal approximation $\hat{S}(t)$ in the Adaptive Spiking Neuron computes a ReLU function: plotted in Figure \ref{fig:ReLuVariance}a is both the firing rate (dashed) and the mean and standard deviation of the signal approximation $\hat{S}(t)$ (solid) for increasing signal values $S$, for two different ratios of $\vartheta_0$  and $m_f$. While the firing rate saturates, the approximation $\hat{S}(t)$ remains linearly growing with increasing $S$, albeit with increasing variance as the number of spikes used to encode the signal remains the same. 

Since the ratio of the baseline threshold $\vartheta_0$ and the multiplicative factor $m_f$ determines the saturating firing rate, this ratio also determines the precision of the encoding. The inverse relationship between saturating firing rate and coding precision is plotted in Figure \ref{fig:ReLuVariance}b,c for 5 different values of $m_f/\vartheta_0$. We observe that the standard deviation linearly increases with signal magnitude, and inversely relates to the saturating firing rate.

\begin{figure}[t]
  \centering
  \includegraphics[width=0.99\linewidth]{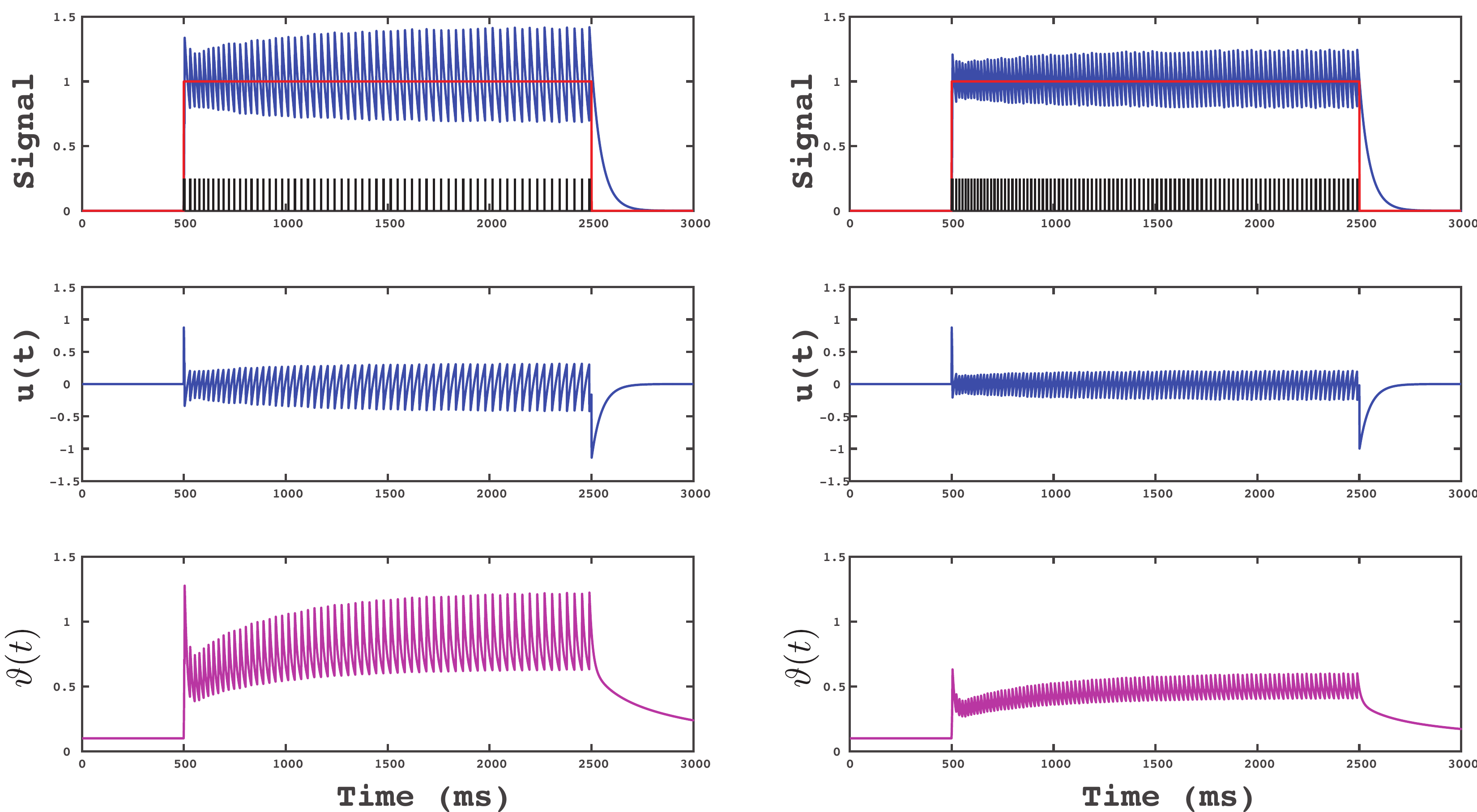}
 \caption{\small Encoding of two fixed size step functions for $S(t)$, illustrating the decreasing variance of the signal approximation $\hat{S}(t)$ for increasing firing-rates. Parameters: $m_f = 1\vartheta_0$ (left) and $m_f = 0.1\vartheta_0$ (right) for $\vartheta_0=0.1$.}
\label{fig:sigApprox}
\end{figure}

In Figure \ref{fig:sigApprox}, we illustrate signal encoding with the ASN with more or less spikes. In the top row we plot the encoding of a step-function $S(t)$ (red) with a sum of adaptive kernels, $\hat{S}(t)$ (blue). The black dashes denote the spikes: the variance of $\hat{S}(t)$ decreases when more spikes are used. In the middle row, the membrane potential $u(t)$ is plotted for both cases, and in the bottom row the dynamical threshold $\vartheta(t)$. As can be seen, a lower firing  rate is achieved by a higher average threshold and correspondingly larger refractory reset $\eta(t)$. Note that the post-synaptic neuron observes the signal $\hat{S}$ smoothed by the filter $\phi(t)$ (which is a better approximation of the step-function at the expense of additional delay).

\begin{wrapfigure}{r}{0.52\textwidth}
\vspace{-0.4cm}
\begin{center}
\includegraphics[width=0.5\textwidth]{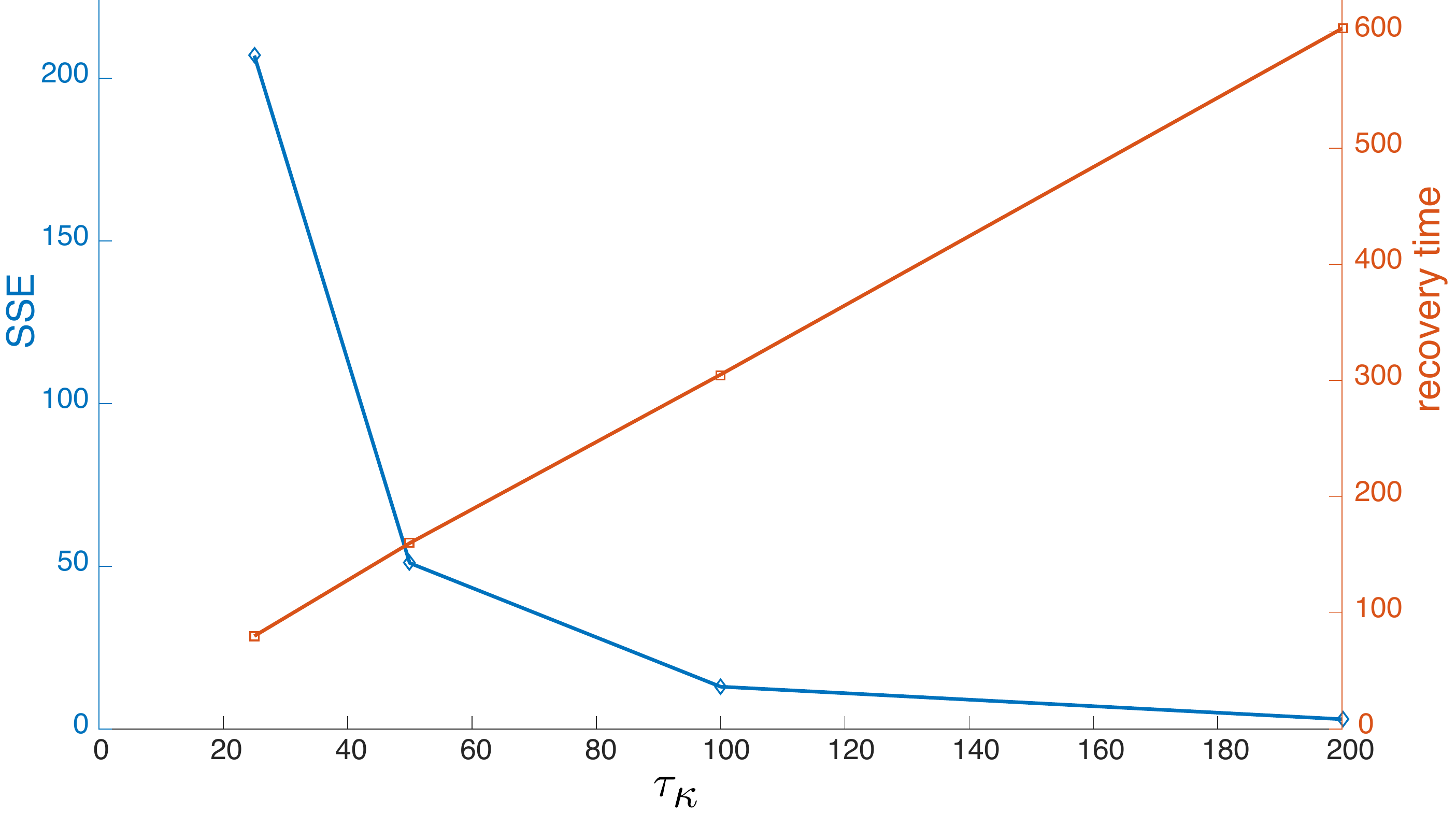}
\end{center}
\vspace{-0.2cm}
\caption{\small Error and responsiveness when encoding a step-function with different $\eta(t)$ (or EPSP) time-constants $\tau_{\kappa}$. Left axis: sum-squared error. Right axis: responsiveness when switching back.}
\label{fig:taus}
\end{wrapfigure}

The time-constant of the refractory response $\eta(t)$ is determined by $\tau_{\kappa}$: the value of this constant determines how much ``future'' signal each spike transmits. To encode step-functions as in figure \ref{fig:sigApprox}, a decay constant that better matches the temporal correlation in the approximated signal will yield a better approximation. For a step-function, this effect is plotted in figure \ref{fig:taus}. Shown is the error (SSE) approximating a 1 second segment of a step-function with a fixed firing rate (35Hz) for various values of $\tau_{\kappa}$. Increasing $\tau_{\kappa}$ strongly reduces the SSE (blue line, left axis). The lower SSE however comes at the expense of responsiveness: when the step-function steps back to 0, it takes longer before the approximation correctly matches the new, lower value. Plotted also (orange line, right axis) is the time it takes before the signal approximation is below 0.05 after stepping down.

\paragraph{Implementation} 
In the examples and in our network implementations, we use time constants that are roughly of the order of the corresponding values in biological spiking neurons, such as time constants of PSCs, membrane time-constant and refractory response kernels, to obtain plausible firing rates for active neurons (1-100Hz). We use a time constant of $\tau_{\kappa} = 50$ms for the exponential decay of the $\kappa$ kernel. The $\gamma$ kernel was approximated as a single decaying exponential $\gamma(t) = e(-t/\tau_{\gamma})$, with time constant $\tau_{\gamma} =15$ms and weight $\gamma=1$.
We use a time constant of $\tau_{smooth} = 2.5$ms for the signal reconstructing exponential smoothing filter $\phi(t)$ in all ASN units except for the output neurons. In the output units activity was filtered with an exponential filter with a longer time constant of $\tau_{rout} = 10$ms, to compare activations between outputs for classification purposes. The simulations are computed with time-steps of size 1ms.

%% file: methodsondataset.tex
\section{Adaptive Spiking Neural Networks (ASNN) vs Artificial Neural Networks}
\label{sec:asnnvsann}
\begin{figure}[t]
\begin{center}
\includegraphics[width=1.0\linewidth]{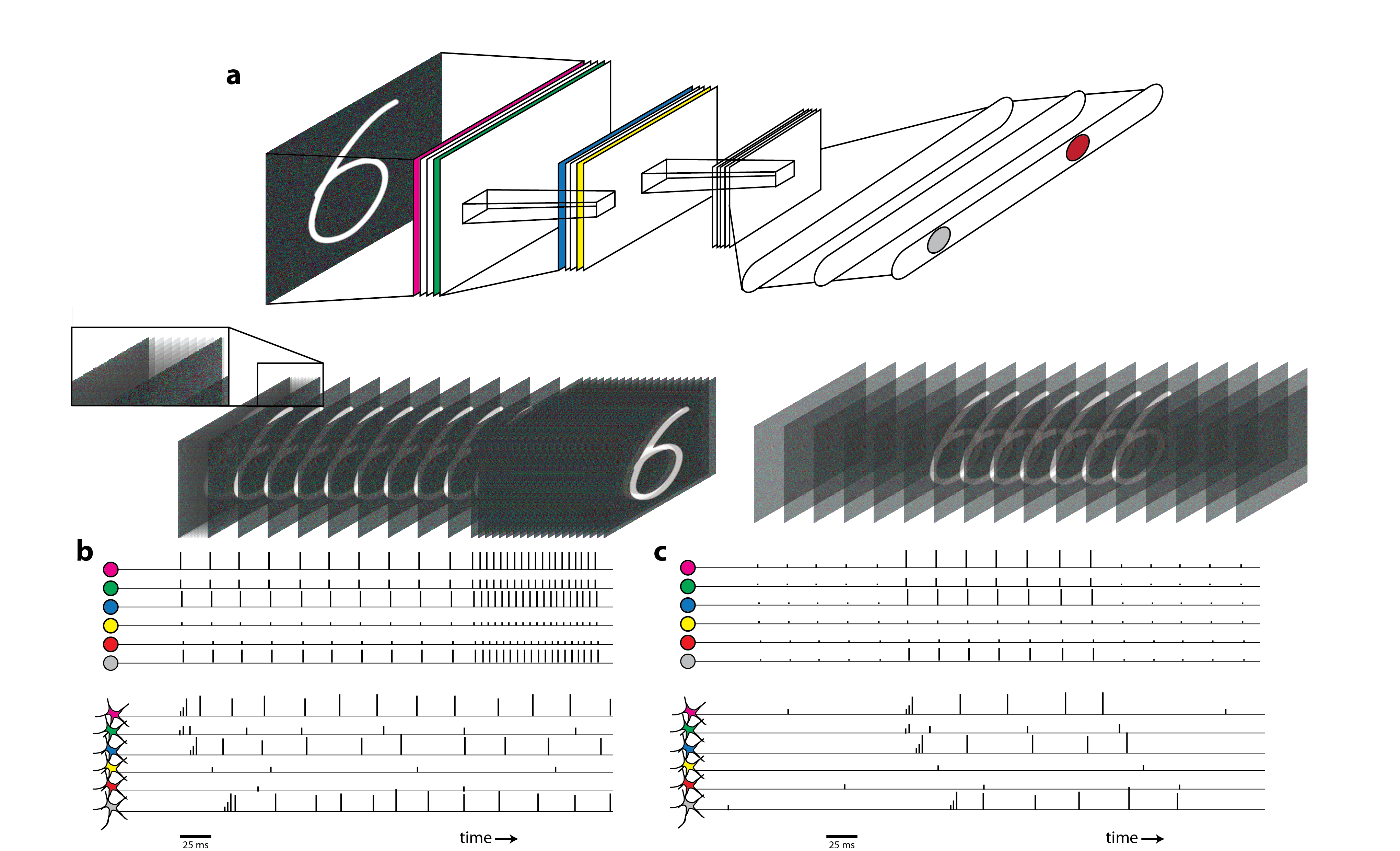}
\end{center}
\caption{\small (a) deep convolutional neural network. (b) ANN vs ASNN classification. The ANN is computed for every frame, for the ASNN the neuron are updated at a fine resolution (inset), but network activity is asynchronous and sparse. Right part of the sequence: increasing the frame-rate increases ANN computations and not ASNN. (c) Flanked noise classification. The ANN computes at a fixed frame-rate, also for noise input that activates feature neurons only slightly. 
For the ASNN, the input neurons rarely cross threshold and the network firing rate is very low for noise; spikes are only emitted when frames with features are presented.}
\label{fig:asnnvsann}
\end{figure}

We implement Adaptive Spiking Neural Networks where the units are comprised of the ASNs described above. Inherently, the ASNNs compute over time-continuous input signals; most straightforward and standard applications of deep neural networks are concerned with classification tasks, such as determining the digit in an image (Figure \ref{fig:asnnvsann}a). To compare classification performance between a standard ANN and an SNN, Diehl et al \cite{diehlIJCNN2015} presented the image for certain time-period to the network (typically 500ms), and recorded from the output neurons to determine the classification. The image is thus taken as input to the network for every time-step in the SNN, which may be as small as 1ms (1000Hz) (illustrated in the inset in Figure \ref{fig:asnnvsann}b). 

Since our ASNs communicate analog valued spikes rather than binary spikes, the question is how the classification problem thus phrased compares to a standard ANN which also communicates with analog values. For an image, an ANN can obviously compute the classification in one go, essentially using just one "analog spike". We argue that the correct comparison between SNNs, ASNNs and ANNs is to treat the classification problem as a time-continuous problem. While the stimulus is present the network has to compute classifications. For both SNNs and ASNNs this is inherent to the operation of the network, while an ANN would need to sample the input at a certain frame-rate. This is illustrated in Figure \ref{fig:asnnvsann}b: the ANN computes the classification for each frame for the entire network, and the computational complexity scales linearly with the frame-rate (illustrated in the right part of Figure \ref{fig:asnnvsann}b). In contrast, the SNN and ASNN implement an asynchronous model of ongoing neural computation where the neurons are updated each small timestep (1ms), and communication between neurons is both localised (to active neurons), and a function of desired neural coding precision rather than frame-rate. Another benefit of the ASNN implementation is illustrated in Figure \ref{fig:asnnvsann}c: when no features are present in the frame, the spiking neural network does not generate spikes, or only very sparingly, whereas the ANN still computes the entire network every frame. The downside of asynchronous neural computation is that there is an inherent latency between input presentation and output: in each layer, the ASN applies an averaging filter to the spike-triggered input currents it receives. Conforming to biological data, we set the filter's timeconstant as 2.5ms, and used a 10ms averaging filter in the output neurons. 

\begin{figure}[t]
\begin{center}
\includegraphics[width=1.0\linewidth]{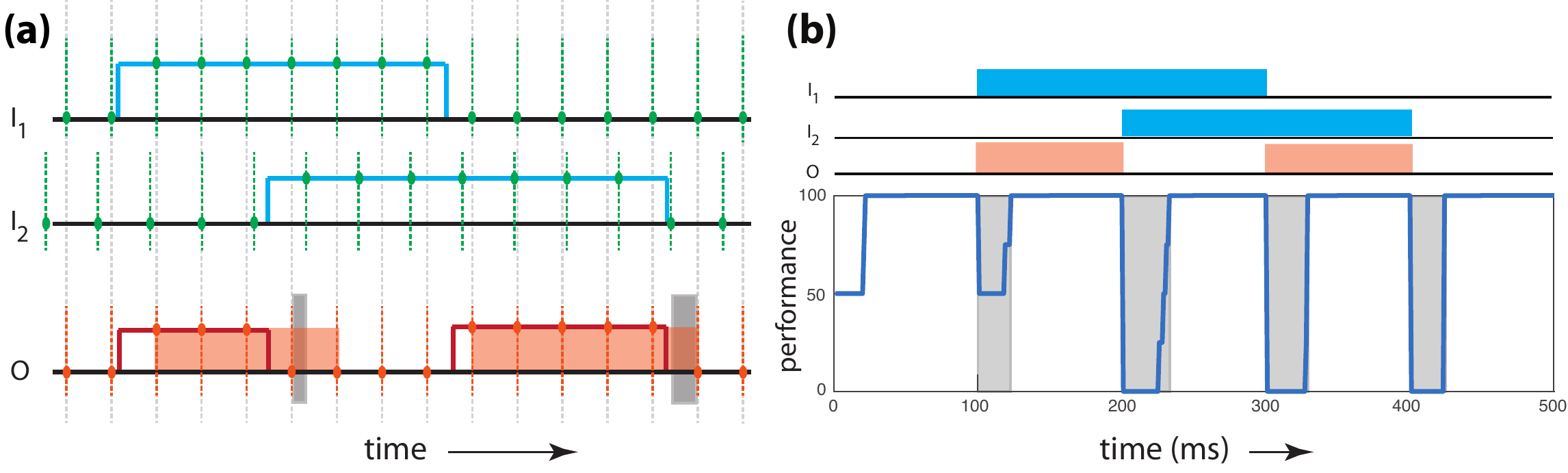}
\end{center}
\vspace{-0.4cm}
\caption{\small Asynchronously computing XOR: (a) illustration with inputs arriving asynchronously (dotted green lines), and XOR computed synchronously with the top (fastest) input-rate. Due to the synchronous nature of computing, additional error are made, like the shaded areas in the bottom figure. Processing the input asynchronously at their respective sample rates, the right shaded area would be avoided. (b) Asynchronous processing of XOR in a 2-5-1 ASNN network capable of computing XOR with about 15Hz average firing rate and neurons using $\tau_{\kappa}=25ms$. Novel input is processed at the update rate of the neurons (1ms); the delay in classification when patterns switch is now determined by $\tau_{\kappa}$ (shaded areas).}
\label{fig:snnxor}
\end{figure}

Asynchronous neural computation offers benefits both for computing and for processing sensory motor data: with neural updating and network updating decoupled, sensory inputs (and actuator outputs) can be sampled at the high neural update frequency. This avoids the well known problem of synchronized processing \cite{olson2010passive}; the ASNN however cannot respond much faster to changing inputs than the $\tau_{\kappa}$ time-constant. This is illustrated in figure \ref{fig:snnxor} for the simple problem of streaming XOR: the network, using about a 15Hz average firing rate, computes XOR from the two inputs. The bottom panel shows performance, and demonstrates that the network is still capable of responding faster to changes in input ($\approx$25ms) than a correspondingly synchronous sample rate. 

\paragraph{Computational Complexity}
An examination of the computational cost and bandwidth requirements demonstrates the mixed ANN and SNN properties of the ASNN. In Table \ref{tab:ccost}, these cost are specified. The ASNN shares the firing rate dependent network bandwidth cost with the SNN, but at an ANN-like cost per spike, and network delay is determined by the spike-decay timeconstant $\tau_{\kappa}$, (presumably) the same as in the SNN (not demonstrated in the literature). Since spike-impact is computed as the product of spike-height and connection weight, the ASNN shares the ANN's cost in terms of multiplications per spike/update, and the neuron update cost of the ASNN scales as an SNN.

\begin{table}[t]
  \caption{\small Computational Cost. $C$: number of connections, $P$: pulse precision, $H_a$: ANN update frequency, $O$: addressing overhead, $F_s$: SNN firing rate,  $F_p$: ASNN average firing rate, $L$: network depth (layers), $U_s$: update frequency of SNN, $U_p$: update frequency of ASNN, $c$: a constant.}
  \label{tab:ccost}
  \centering
  \begin{tabular}{lrrr}
    \toprule
     & ANN & SNN & ASNN \\
    \midrule
    Network bandwidth    &  $C \cdot [P + O]  \cdot H_a $   & $C \cdot O \cdot F_s$ & $C  \cdot [P+O] \cdot F_p$ \\
    Network delay  &  $1/H_a$   & $ (?) \tau_{\kappa} + c\cdot L $ & $ \propto \tau_{\kappa} + c\cdot L$ \\
    Network multiplications & $C \cdot P \cdot H_a$ & -  & $C  \cdot P \cdot F_p$ \\
    Neuron multiplications & $H_a \cdot f(\text{ReLU})$ & $U_s \cdot f(\text{ReLU})$  & $U_p [ 3 + f(\text{threshold})] $ \\
    \bottomrule
  \end{tabular}
\end{table}

This analysis ignores the fact that spikes in the ASNN (and SNN) are heavily localized to a subset of neurons: many neuron are silent while a few are active. Sparse and localised communication potentially offers a benefit to deep neural networks, as densely connected neural networks tend to be limited by the bandwidth required to read and write the appropriate weights from memory \cite{slazynski2012streaming}. Thus reasoned, for an ASNN that incurs a 100ms delay to compete in terms of bandwidth used with an ANN, it can use at most a firing rate of 10Hz on average per neuron, since an ANN sampled with 10Hz would achieve the same worst case delay. This ignores the benefit of the ASNN being able to process in principle a 1000Hz frame-rate. The exact benefit of sparse activity depends on the degree of sparseness and the degree to which parallel hardware can exploit sparseness.
 
\section{Feedforward and Deep Convolutional ASNNs}
We tested networks of multiplicative Adaptive Spiking 
Neurons (ASN) described above, both in fully connected Feed-Forward Neural Networks (FFNNs) and in a Convolutional Neural Network (CNN) \cite{Lecun:1998hy}. These architectures were first trained on standard datasets -- IRIS, SONAR, and MNIST -- with standard ANNs comprised of rectified linear (ReLU) neurons. The corresponding spiking neural networks were created by using the same weights and network connectivity as the trained architectures (similar to \cite{diehlIJCNN2015}), and replacing the ReLU neurons with ASN units. 

We selected well-known benchmark datasets of increasing complexity to demonstrate the robustness of the presented approach. The IRIS dataset is a classical non-linearly separable ``toy'' dataset containing 3 classes -- 3 types of plants -- with 50 instances each, to be classified from 4 input attributes. Similarly, the SONAR dataset \cite{Gorman:1988dx} contains 208 entries of sonar signals divided in 60 energy measurements in a particular frequency band, to be classified in metal cylinder or simple rocks classes. Lastly, we use the MNIST dataset \cite{Lecun:1998hy}, which has been a standard testbed for novel image classification methods. It is composed of 60000 entries of handwritten digits for the training set and 10000 entries for the validation set. 

To carry out classification, for each instance the input neurons receive input current $I(t)$ corresponding to the respective feature values, for a simulation duration of 500ms. During this period, input neurons generate spikes that are instantaneously transmitted to the next layer. There, the corresponding weighted PSCs are added to the membrane potential $u(t)$ through the smoothing filter $\phi(t)$; note that the smoothing filter effectively causes a delay in signal transmission of order $\tau_{smooth}$ per layer. This process is repeated for each successive layer in the network. Output values used for classification are computed as internal current $I(t)$ in the output neurons, smoothed with longer time constant $\tau_{rout}$ for stable performance. At every 1ms timestep $t$ of the simulation, classification performance is computed over all instances of the respective dataset from the outputs $I(t)$ at that timestep $t$. 

\paragraph{Feed-Forward Neural Networks}
We trained fully connected FFNNs using dropout \cite{Srivastava:2014ww} to approximately match performance with state-of-the-art. We trained a four layer FFNN of size $[4-30-30-3]$ on the classical IRIS dataset with a dropout rate of $0.5$, learning rate of $0.1$, for $800$ epochs. We used half of the dataset for training, and we obtained $97.33\%$ on the validation set. For the SONAR dataset, we trained a four layer FFNN of size $[60-50-50-2]$, using the training set division reported in \cite{Gorman:1988dx} for the angle-dependent experiment. We used a dropout rate of $0.5$, learning rate of $0.2$, and we trained for $1000$ epochs to obtain $88.46\%$ accuracy on the validation set. For the MNIST dataset, we used the trained network reported in \cite{diehlIJCNN2015} to directly compare with the method there. In \cite{diehlIJCNN2015}, the authors trained a  $[784-1200-1200-10]$ network, with a dropout rate of $0.5$, learning rate of $1$ and momentum of $0.5$. With this network, we obtained $98.84\%$ accuracy on the MNIST validation set (code and trained network were available online\footnotemark[1]\footnotetext[1]{\label{first}\url{http://github.com/dannyneil/spiking_relu_conversion}}) using a modified version of the DeepLearnToolbox\footnotemark[2] \cite{IMM2012-06284}. \footnotetext[2]{\url{https://github.com/rasmusbergpalm/DeepLearnToolbox}}As in \cite{diehlIJCNN2015}, for all datasets the input values were scaled to the range $[0,1]$. We refer to the FFNNs that use ASN ReLU neurons as Feed-Forward Adaptive Spiking Neural Networks (FF-ASNN).

\paragraph{Convolutional Neural Networks}
CNNs have become a standard tool for image classification tasks \cite{Lecun:1998hy}, and they generally outperform classical FFNNs. In \cite{diehlIJCNN2015} a competitive ReLU CNN implementation for MNIST was presented: we apply the ASN network to this architecture and compare our results to those obtained in \cite{diehlIJCNN2015}. The pre-trained network consists of a $[28\times28-12c5-2s-64c5-2s-10o]$ CNN, where $28\times28$ corresponds to the input image size, $N$ $c$ $K$ are the N-convolutional kernels of size $K$, $M$ $s$ $J$ are the M-averaging pooling filters of size $J$, and $o$ is the size of the output layer; note that this network is available online\footnotemark[1]. Neurons in each of these layers use the ReLU activation function, and we can again map the ANN directly to our ASNN by substituting each ReLU neuron with the Adaptive Spiking Neuron. We refer to the CNNs equipped spiking neurons as Convolutional Adaptive Spiking Neural Networks (C-ASNN).

%% file: resultsffnn.tex
\section{Results}

\begin{figure}[t]
\begin{center}
\includegraphics[width=0.999\linewidth]{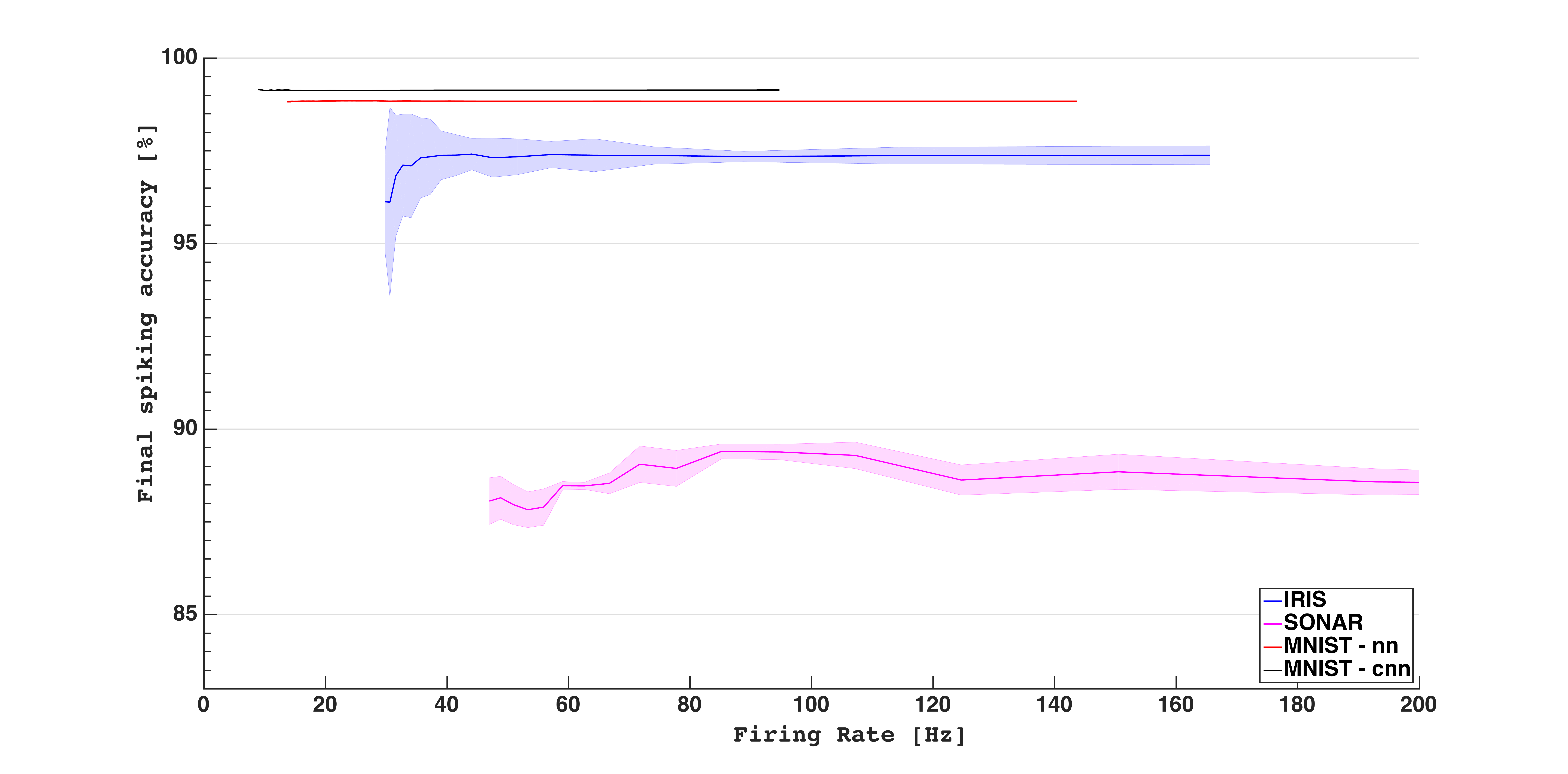}
\end{center}
\caption{\small Classification performance on IRIS, SONAR, MNIST (MNIST-nn for FF-ASNN and MNIST-cnn for C-ASNN) for various average firing rates. Dashed: performance of original ANN. }
\label{fig:fratePerf}
\vspace{-0.25cm}
\end{figure}

For all three datasets and the corresponding four ReLU architectures, we computed the ANN performance and compared that to the ASNN performance. Figure \ref{fig:fratePerf} shows classification performance obtained for IRIS, SONAR and MNIST by the various ASNNs as a function of average firing rate in the network (and hence neural coding precision) during classification, obtained by varying the ratio of $m_f$ and $\vartheta_0$. We find that for all benchmarks we achieve performance with the ASNN identical to that of the corresponding ANN once a certain minimum firing rate is used, corresponding to the minimal required neural coding precision. The networks that classify the IRIS and SONAR benchmarks require fairly high firing rates compared to the two MNIST architectures. Since the former architectures are comprised of far fewer neurons as compared to the MNIST networks, this suggests that in such smaller networks the coding precision needs to be quite high. 

The different firing rate regimes were obtained by  varying the multiplicative factor $m_f$ as a function of $\vartheta_0$, between $0.1\vartheta_0$ and $3\vartheta_0$, with $\vartheta_0=0.0128$ for the IRIS dataset, in 30 different simulations. The threshold $\vartheta_0=0.0128$ was selected such that the smallest positive input values in the training set were still encoded. For SONAR, we carried out simulations with $m_f$ ranging between $0.1\vartheta_0$ and $3\vartheta_0$, using $\vartheta_0=1e^{-4}$. For the MNIST dataset we simulated both an FF-ASNN and C-ASNN architecture. For the FF-ASNN we carried out $35$ simulations with $m_f$ ranging between $0.1\vartheta_0$ and $3.5\vartheta_0$, using $\vartheta_0=3.9e{-3}$. For the MNIST networks, compared to the IRIS and SONAR networks, we find that performance is stable over a much greater range of firing rates. For each simulation we computed the time to which $101\%$ of the minimum classification error is reached (Matching Time, MT), e.g., for MNIST-cnn this is when the performance exceeds $99.13\%$. Given parameters $\vartheta_0$ and $m_f$, we considered the ASNN network as having performance identical to the corresponding ANN if, in the time-window from MT to the end of the simulation ($500ms$), the performance stays, on average, above the $101\%$-error threshold. The variance is computed over the same time-window, while the firing-rate is computed in a time-window of $100ms$ at the end of the simulation. At low firing-rates, the ANN performance is exceeded for some ranges by chance; the high neural coding precision for higher firing rates results in more stable performance, as can be seen in the low variance of the performance on the right part of Figure \ref{fig:fratePerf}.

For all four ASNNs, we noted both the required minimum firing rate (as set through the ratio of $m_f$ and $\vartheta_0$) to reach the $101\%$-error threshold, and the corresponding simulation time when this performance is first reached. We refer to these values as the Matching Firing Rate (FR) and the Matching Time (MT), and the results are shown in Table \ref{tab:lowFR} in the column {\bf Lowest FR}. For MNIST, we find that the response time for the FF-ASNN is substantially faster as compared to the C-ASNN. This is likely caused by the fact that the C-ASNN is a deeper network. Additionally, we determined the lowest Matching Time and corresponding Firing Rate (Table \ref{tab:lowFR} in the column {\bf Lowest MT}). We see that for the large MNIST networks, Matching Time improves substantially at limited cost in terms of FR. In general, we find that the Matching Time increases with lower firing rates (not shown). 

\begin{table}[t]
  \caption{\small Performance($\%$), Matching Firing Rate (FR) (Hz) and Matching Time (MT) (ms). Performance is compared to the corresponding standard ANN and the Poisson SNN (P-SNN) in \cite{diehlIJCNN2015}.}
  \label{tab:lowFR}
  \centering
  \begin{tabular}{lrrrrr|rl}
    \toprule
    DataSet & ANN & P-SNN & {\bf ASNN} & \multicolumn{2}{c}{Lowest FR}   & \multicolumn{2}{c}{Lowest MT} \\
            \cmidrule{5-8}                   
         & & P($\%$)@FR  & P($\%$) &FR &MT  &FR & MT\\
    \midrule
    IRIS    & 97.33 & - & {\bf 97.33} & 36 & 107         & 41.4 & 46\\
    SONAR   & 88.46 & - & {\bf 88.46} & 59.7 &   80            & 77.1 &   71  \\
    MNIST-nn& 98.84 & 98.64@1000& {\bf 98.84} & 14.6 & 15       & 17.3 & 12 \\
    MNIST-cnn& 99.14& 99.12@1000& {\bf 99.14} & 8.6 & 87       & 10 & 8.9 \\
    \bottomrule
  \end{tabular}
\end{table}

\input{switching}

%% file: switching.tex
\paragraph{Switching}

\begin{figure}[h]
\begin{center}
\includegraphics[width=0.999\linewidth]{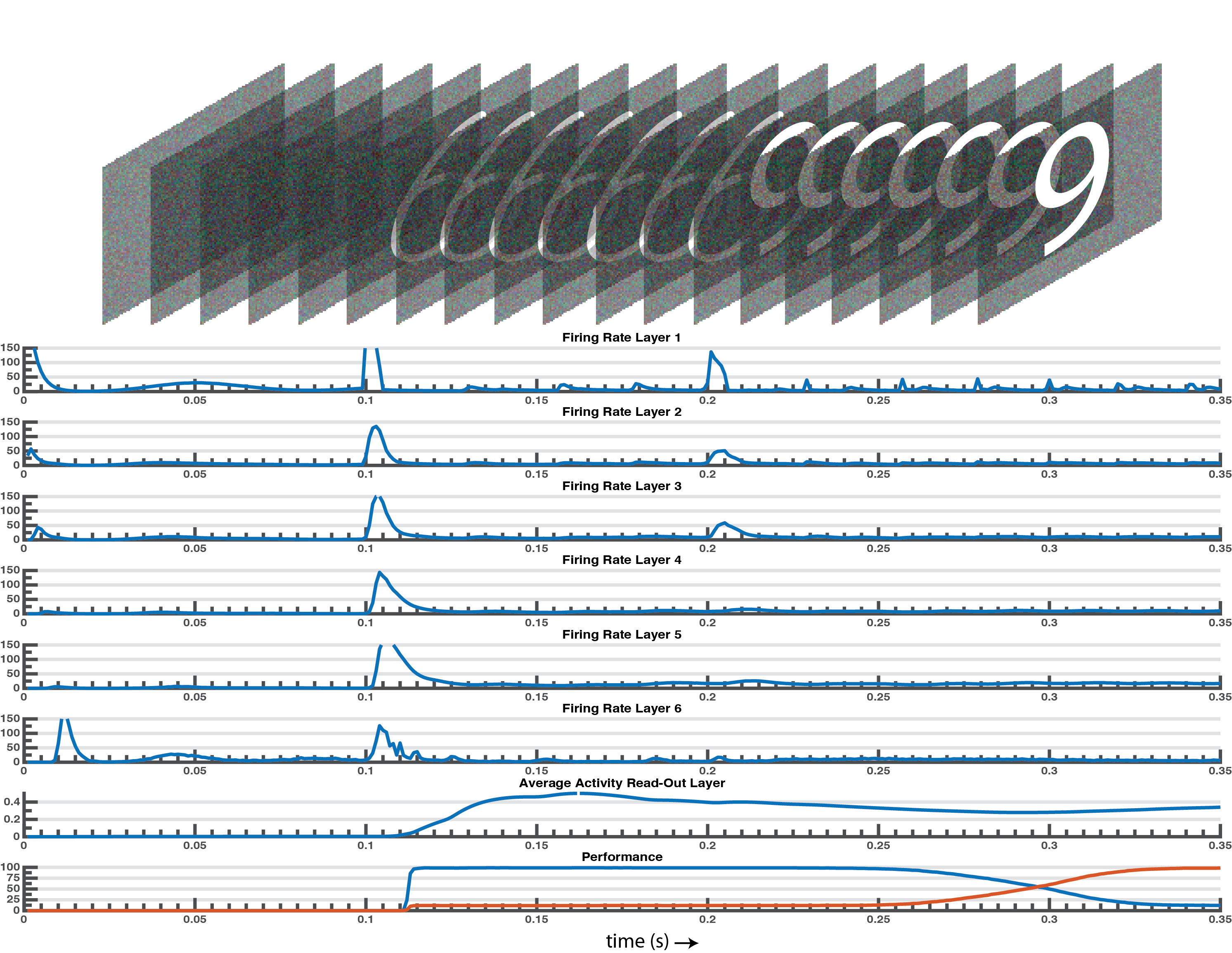}
\end{center}
\caption{\small Switching example with C-ASNN: random, 1000 digits, permutation of the 1000 digits. Top: an example of the switching images provided to the network. Middle, rows 1-6: the firing rate of the network's 5 layers plus the read-out layer. Middle, row 7: the average activity of the read-out layer, computed by filtering the internal state of the neurons. Note that, during the noise presentation, although a firing activity in the read-out layer is present, the internal state is null. The rise of the average activity signals that a classification is made. Bottom: the classification performance through time showing the switch between the two test sets. We set a threshold on the read-out activity at $0.3$ and we only computed performance if any of the output neurons exhibited activity above this threshold.}
\label{fig:switchingFigure}
\vspace{-0.25cm}
\end{figure}

As in \cite{diehlIJCNN2015}, we computed the Matching Time to determine that time that input needs to be presented to the network before the output classification reaches ANN performance ($101\%$ of the minimum classification error). A more general streaming setting however is one where one stimulus is presented, followed by another stimulus. We illustrate this case in figure \ref{fig:switchingFigure}: first, white noise is presented to the network for $100ms$, followed by the presentation of a digit, which after $100ms$ is then switched to another digit. Shown is the average activation in each layer of the MNIST-cnn ($\vartheta_0=3.9e{-3}$, $m_f = 3\vartheta_0$) for 1000 random stimulus switches, as well as the average activation $S(t)$ in the output neurons and the classification performance. White noise has been reproduced by presenting a (different) Gaussian-noise sampled image with $\mu = 0$ and  $\sigma = 0.5\vartheta_0$, at each $ms$ frame. We see that noise only stimulates the first layer, and fails to substantially activate subsequent layers. Once the first actual digit is presented, the network rapidly and correctly recognizes this digit. After $200ms$ the permuted images are presented: the classification performance for the new dataset reaches the $101\%$-error threshold after a switching time of $ST = 186ms$. This switch from one digit to another is determined by -- substantially longer -- recovery time due to $\tau_{\kappa}$. Switching time can be improved by decreasing $\tau_{\kappa}$, but at the expense of an increase in firing rate.

%% file: discuss.tex
\section{Discussion}

We constructed an adaptive spiking neural network using fast adapting spiking neurons. Spiking neuron models like the ASN presented here capture many important adaptation phenomena in real neurons, and by coupling the synaptic plasticity model, we ensure that downstream neurons appropriately account for adaptation in presynaptic neurons. Thus, it is a prediction of this work that a tight coupling exists between neural adaptation and synaptic plasticity.  At the same time, we demonstrated that the resulting neural network model can replace a standard ANN in a one-to-one manner, without loss of performance, while using an asynchronous and sparse model of spike-based neural computation. As such, the presented ASNN can be considered as a novel paradigm for neural coding with spiking neurons, with an almost direct correspondence to biological spiking neurons.

In particular, we show that the proposed ASNNs can carry out neural computation with performance identical to the corresponding ANN for a number of classical benchmark datasets of increasing network size and complexity. Compared to an otherwise identical SNN that uses Poisson spiking neurons the presented approach has better or identical performance while using a much lower firing rate in the network. Additionally, due to the large dynamic range of the ASNs, no reweighting or normalization of the network was necessary: the ASNs function as drop-in spiking neuron replacements for the ReLU neurons in the standard ANNs. Effectively, the ASN computes using {\em adaptive} Asynchronous-Sigma-Delta Pulse Modulation, which is necessary because -- unlike electrical circuit signals -- the signals inside a neural network with ReLU neurons are not bounded to some fixed dynamic range. 

Compared to classical ANNs, the computations of the ASNNs are asynchronous, event driven and sparse. To truly exploit the efficiency of sparsely active asynchronous spiking neural networks, efficient GPU or ASIC implementations need to be created. Current CNN implementations are heavily optimized for carrying out convolutions on GPUs, an operation which closely fits the GPUs parallel architecture. For sparsely active neural networks, where most neurons are not active at any given time-step, novel approaches need to be developed: since typically for any stimulus only a subset of neurons is active, fast caching methods are likely to hold promise. As most network of spiking neurons, the reduction in communication between the neurons is traded against more complex dynamics in the neuron; since there are typically orders of magnitude fewer neurons than connections, this tradeoff can be worthwhile provided that the neuron model requires limited memory and computation. The ASN model presented here can be computed with only a few variables (principally the components of the $\gamma$ and $\eta$ kernels), which when formulated as simple dynamical systems can be computed in a memory-less fashion, without tracking previous spike-times. 

Compared to non-adaptive networks, adapting neurons effectively use analog spikes: each spike is associated with a refractory kernel of different height. In principle, the analog value of a spike can be reconstructed at the postsynaptic neuron from just the time since the previous spike, but at considerable computational expense. Compared to standard (analog) ANNs, the ASNNs compute in an asynchronous and localized manner: input information can be presented to the network at the precision with which neurons are updated, while the rate of information exchange in the network is determined by the neural coding precision required for classification. The network can thus process for instance 1000Hz input frames when neural updates are carried out with 1ms timesteps: in this manner, new input can be processed almost immediately -- albeit with the delay incurred in the consecutive layers. The neural activity is also localized, in that only a subset of neurons is really activated, emitting many spikes, and most neurons are silent or only very sparsely active. Since bandwidth, as used for reading weights from memory, is typically the limiting factor when computing an ANN, the sparse and localized neural computation offers a potentially more efficient way of time-continuous neural computing. 

The networks presented here are specific in that they comprise of fairly straightforward neural networks without additions like pooling layers. Many current state-of-the-art deep neural networks however comprise of a number of different layers which do not adapt to networks of ReLU neurons in a straightforward manner: for operations like softmax or layer-wise normalisation, specific SNN variations need to be developed.